\documentclass[letterpaper, 10 pt, conference]{ieeeconf}

\IEEEoverridecommandlockouts

\usepackage{graphics}
\usepackage{epsfig}
\usepackage{mathptmx}
\usepackage{times}
\usepackage{amsmath}
\usepackage{amssymb}
\usepackage{booktabs}
\usepackage{graphicx}
\usepackage{capt-of}
\usepackage{xcolor}

\makeatletter
\long\def\@makecaption#1#2{%
\ifx\@captype\@IEEEtablestring%
\begin{center}{\footnotesize #1}\\{\footnotesize\scshape #2\par}\end{center}%
\@IEEEtablecaptionsepspace%
\else
\@IEEEfigurecaptionsepspace%
\setbox\@tempboxa\hbox{\footnotesize #1.~~ #2}%
\ifdim \wd\@tempboxa >\hsize%
\setbox\@tempboxa\hbox{\footnotesize #1.~~ }%
\parbox[t]{\hsize}{\footnotesize \noindent\unhbox\@tempboxa#2}%
\else%
\ifcenterfigcaptions \hbox to\hsize{\footnotesize\hfil\box\@tempboxa\hfil}%
\else \hbox to\hsize{\footnotesize\box\@tempboxa\hfil}%
\fi\fi\fi}
\makeatother

\title{\LARGE \bf
DexStitch: Learning and Composing Dexterous Skills from Human Videos via Contact-Aware Robotic Inpainting
}

\author{Byeonggyeol Choi$^{1,*}$, Seongmin Jung$^{1,2,*}$, Unghui Nam$^{1,*}$, Youngbo Song$^{1,*}$, Sunhong Min$^{3,*}$, \\
MinHyung Lee$^{3}$, Byonghyo Shim$^{1}$, and Yongseok Lee$^{3,\dagger}$%
\thanks{$^{*}$These authors contributed equally to this work.}%
\thanks{$^{\dagger}$Corresponding author.}%
\\[0.6em]
{\normalsize $^{1}$Seoul National University \quad
$^{2}$Robotics Lab, Hyundai Motor Company \quad
$^{3}$DGIST}%
}

\IEEEaftertitletext{%
  \vspace{-\baselineskip}%
  \begin{minipage}{\textwidth}
    \centering
    \includegraphics[width=\textwidth]{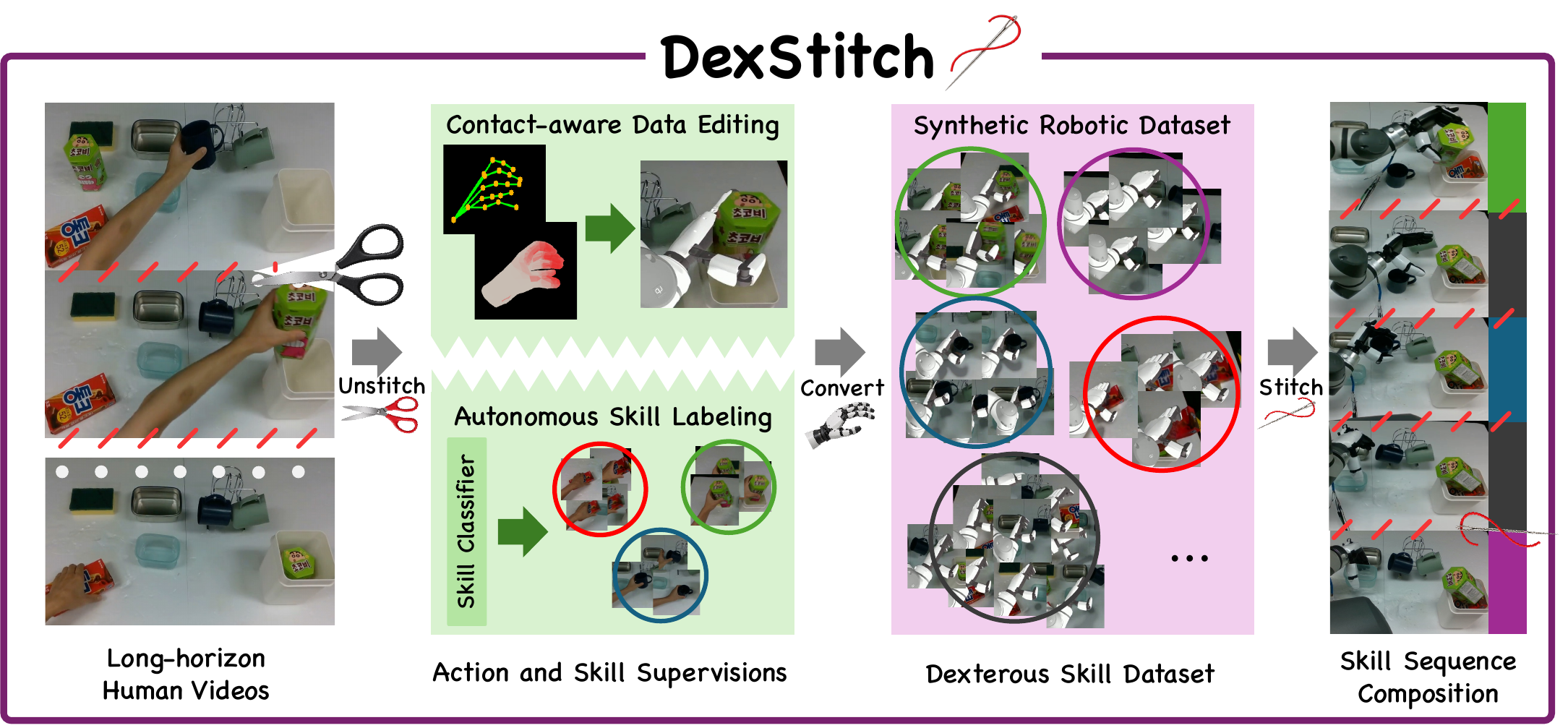}
    \captionof{figure}{
    \textbf{System overview.}
    We propose \textbf{DexStitch}, a system that cuts human videos into skills and actions, converts them into a dexterous robot dataset, and stitches the learned skills back together in a sequential order.
    }
    \label{fig:overview}
\end{minipage}
\vspace{-1.0\baselineskip}

  \vspace{1\baselineskip}%
}

\begin{document}

\maketitle
\thispagestyle{empty}
\pagestyle{empty}

\begin{abstract}

A long-horizon manipulation task consists of primitive skills and the transitions connecting them.
Learning skills that can be executed in sequence requires supervision for both, which is costly to collect on a robot.
Human videos offer a scalable source of demonstrations, but transferring them to robots requires extracting robot-compatible observations and actions together with their skill structure.
We present DexStitch, a framework that cuts long-horizon human videos into skills and transitions, converts them into aligned robot observation–action–skill data, and stitches the learned skills into a requested sequence.
A skill classifier first segments the videos into manipulation skills and transitions.
DexStitch then converts the human demonstrations into robot observations and actions while retaining their corresponding skill labels.
During conversion, per-finger human–object contacts guide dexterous hand retargeting and robotic inpainting, preserving contact geometry and finger–object occlusion ordering.
To \textit{stitch} the learned skills into a sequence, a shared skill-conditioned policy is pretrained on the resulting synthetic data and finetuned on a mixture of synthetic data and a small set of single-skill teleoperation episodes.
Conditioning on the current and next skills guides each transition toward the following skill, while predicted skill progression enables closed-loop switching.
On a held-out session, the classifier agrees with manual annotations on 92.4\% of frames.
Across four household skills, DexStitch achieves 80.0\% single-skill success and 77.5\% sequence success.

\end{abstract}

\section{INTRODUCTION}

Long-horizon manipulation tasks consist of primitive skills and the transitions connecting them: a robot tidying a desk picks up objects, opens drawers, and places items inside.
Performing such tasks requires learning individual skills and composing them into a sequence.
Successful composition also requires the motions that connect successive skills, since the state left by one skill may differ from the starting states encountered by the next during training~\cite{lee2019transition}.
The robot must also determine when a transition is complete and the next skill can begin~\cite{chen2023sequential}.
Human videos naturally capture both manipulation skills and the transitions between them, offering a readily available source of demonstrations.
To turn these demonstrations into robot-compatible data, we need to recover both how to act and which skill each action belongs to.
This requires robot-compatible observations and actions that preserve the demonstrated interactions, together with aligned skill annotations that organize the sequential execution.

Collecting the action and skill supervision through robot teleoperation and manual annotation is costly, whereas human videos are easier to record but provide no robot actions and skill labels~\cite{zakka2021xirl}.
Existing approaches address this gap in two ways.
One line learns a shared latent skill space across human and robot embodiments, enabling a policy trained on robot data to follow a human demonstration~\cite{xu2023xskill, kim2025uniskill}.
These methods transfer skill structure through latent representations rather than producing explicitly skill-labeled robot observation–action pairs from human videos.
Another line retargets human motion and replaces the human embodiment with a rendered robot, producing robot-domain demonstrations~\cite{lepert2025phantom, lepert2025masquerade, li2025h2r, liu2026egoengine}.
However, this conversion does not by itself identify the skills and transitions within each trajectory.
Together, these approaches motivate a framework that converts human demonstrations into robot-compatible observations and actions while making their skill structure explicit.

We present \textbf{DexStitch} (Fig.~\ref{fig:overview}), which cuts long-horizon human videos into skills and transitions, converts them to the robot domain, and stitches them back together in a requested order.
A user defines the skill set, including transitions as ordinary skills, and annotates a set of videos frame by frame.
A classifier trained on these annotations segments the remaining videos, agreeing with held-out manual labels of frames.
Because transitions are expensive to cover individually---one for every ordered skill pair---we record long-horizon human videos that contain many transitions at once.

Robot actions are obtained by converting the human videos.
Retargeting to a dexterous hand must preserve both the contacting fingers and their contact points; due to hand-shape differences, matching fingertip positions alone may miss the points touched by the human.
From RGB video alone, without an object mesh or simulator, DexStitch predicts per-finger surface contact and uses it to pull retargeted fingertips onto the contact points.
It also determines which fingers are occluded by the object when inpainting the human and rendering the robot.
Because conversion is frame-wise, the skill labels transfer directly to the synthetic robot data.

Our skill-conditioned policy learns both skills and transitions in two stages.
It is pretrained on the synthetic data from manually and classifier-labeled videos, then finetuned with a small set of teleoperated single-skill episodes.
Thus, transitions are supervised solely by human video.
Conditioning on the current and next skills directs each transition toward the next skill's initial state, while an auxiliary prediction of skill completion advances the sequence.
At test time, the same classifier extracts a skill sequence from a human prompt video, and the robot executes the corresponding skills in order.

We evaluate DexStitch on four household manipulation skills.
We first ask whether the synthetic robot data teaches executable skills, comparing against skill learning and compositing baseline~\cite{kim2025uniskill}.
Ablations remove each data source, the transition skill, and the next-skill conditioning.
We then qualitatively compare the converted videos against Masquerade~\cite{lepert2025masquerade} and a variant without contact-guided occlusion handling.
Finally, we measure how accurately the classifier segments held-out human video against two vision--language models.

Our contributions are threefold.
\begin{itemize}
    \item \textbf{Skill and action supervision from human video.}
    We convert long-horizon human videos into robot-domain observation--action pairs while preserving their frame-level skill annotations, providing supervision for both manipulation skills and the transitions between them.
    Using RGB video alone, our pipeline produces robot-executable training data without object meshes or simulation.

    \item \textbf{Contact-aware human-to-robot conversion.}
    We use per-finger human--object contact as a shared cue for retargeting and visual compositing, preserving multi-finger grasp geometry while resolving finger--object occlusions during robot rendering.

    \item \textbf{Sequential composition of reusable skills.}
    We construct an explicit, reusable repertoire of robot-executable skills, including transitions, from the converted data.
    Our policy conditions on the current and next skills to execute target-aware transitions, and predicts skill completion for closed-loop sequential composition.
\end{itemize}

\section{RELATED WORK}

\subsection{Learning Skill Structure from Human Video}

Long-horizon human demonstrations contain behavioral structure that can be represented as manipulation skills.
Prior work has studied such structure in play and unsegmented robot demonstrations, learning latent plans~\cite{lynch2019play}, reusable skill primitives~\cite{zhu2022buds}, and continually expanding skill libraries~\cite{wan2024lotus}.
More recent approaches extend this perspective to human video.
MimicPlay~\cite{wang2023mimicplay} distills human play into latent plans that guide a robot controller trained on teleoperation.
XSkill~\cite{xu2023xskill} and UniSkill~\cite{kim2025uniskill} learn cross-embodiment latent skills from unlabeled human and robot videos, and sequence them to imitate a human prompt video in one shot.
In both, the policy receives a continuous latent.
XSkill's prototypes cover the motion between manipulations too, but unlabeled, and neither policy is told which skill comes next.
DexStitch takes the opposite choice: a skill set is fixed in advance, with the transition as one of its labels.
Human videos are then automatically segmented into the labels.

\begin{figure*}[t]
\centering
\includegraphics[width=0.95\linewidth]{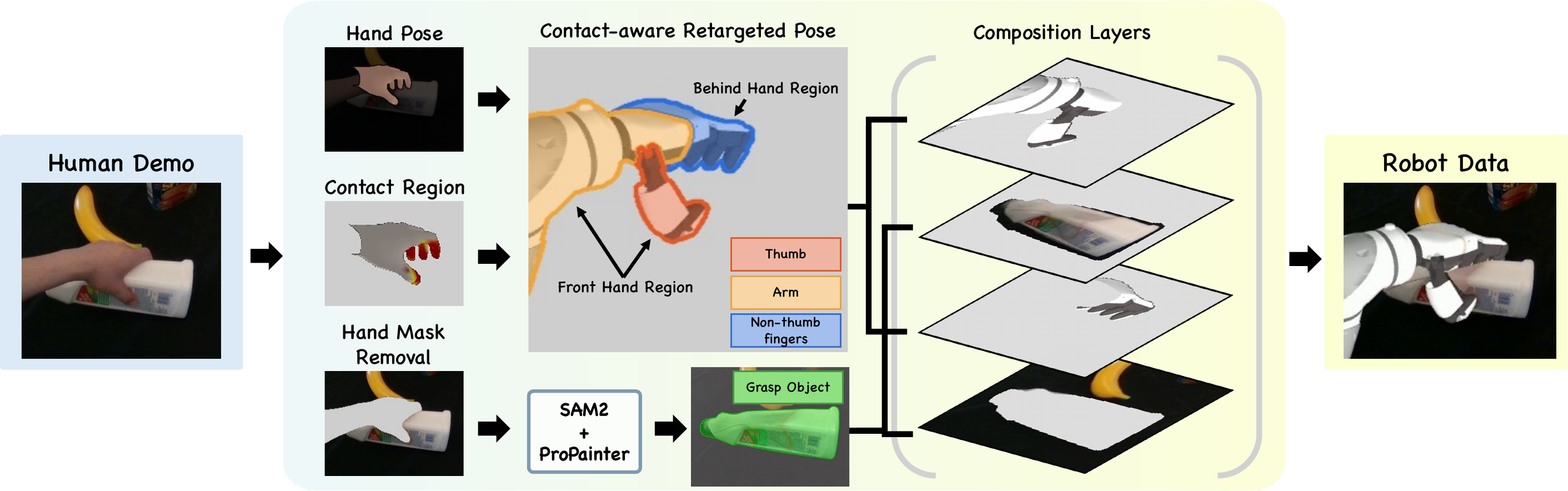}
\caption{
\textbf{Contact-aware Human-Robot Data Conversion Pipeline.}
Given an egocentric human demonstration, DexStitch estimates the hand pose and hand--object contact while removing the human embodiment from the input frames.
The contact information then guides robot retargeting and determines the object--robot occlusion order during compositing.
Together, the retargeted robot and the inpainted scene produce a robot-domain visual sequence.
}
\label{fig:method1}
\end{figure*}

\subsection{Transferring Human Videos to Robot Execution}

Human videos provide abundant manipulation
demonstrations~\cite{grauman2022ego4d,nair2022r3m} but lack embodiment-specific
robot actions and visually differ from observations encountered by the robot.
Motion-retargeting approaches recover robot-compatible actions from human
motion, but the corresponding frames still depict the human
embodiment~\cite{kareer2025egomimic,qiu2025humanoid,yang2025egovla,zheng2026egoscale}.
These methods therefore rely on additional robot data to ground policy learning
in robot-domain observations.
Robotic inpainting instead removes the human embodiment and renders the
retargeted robot into the source
video~\cite{lepert2025phantom,lepert2025masquerade,li2025h2r,liu2026egoengine}.
Phantom~\cite{lepert2025phantom} and EgoEngine~\cite{liu2026egoengine} pair the
edited frames with the retargeted actions and train policies on them without
robot demonstrations, while Masquerade~\cite{lepert2025masquerade} and
H2R~\cite{li2025h2r} use them to pretrain visual representations that robot
demonstrations then ground.
In all four, a demonstration is one trajectory, with no skill or transition
boundaries marked.
DexStitch also trains on the edited frames paired with the retargeted actions,
and adds a small set of teleoperated episodes only in finetuning.

Compositing must also order the rendered robot in depth against the object it
holds, or fingers that should curl behind the object are drawn in front of it.
Phantom~\cite{lepert2025phantom} masks the overlay with depth from an RGB-D
camera, and EgoEngine~\cite{liu2026egoengine} recovers the ordering from a
reconstructed digital twin, at the cost of an object mesh, per-frame depth, and
a physics simulator for the actions.
Contact-aware retargeting methods recover the demonstrated interaction geometry
during motion
transfer~\cite{lin2025dexflow,mandi2025dexmachina,hsieh2025dexman},
but also take their contact from an object model, scanned or reconstructed, and
use it only for the actions.
DexStitch estimates per-finger contact on the hand surface from the RGB video
alone~\cite{jung2025haco}, with neither depth nor an object model, and uses the
same signal both to refine the retargeted hand configuration and to identify
which fingers pass behind the object during compositing.

\subsection{Skill Composition and Transition}

Composing skills in sequence means each skill starts where the previous one ended, which may be a state it never saw in training~\cite{lee2021adversarial}.
Prior work bridges this gap with dedicated transition policies~\cite{lee2019transition}, a learned transition-feasibility function that finetunes and switches sub-policies~\cite{chen2023sequential}, or a motion planner that carries the end-effector between atomic skills served by one multi-task policy~\cite{yang2026lilovla}.
Other approaches compose discrete skills without a transition module: SDP~\cite{gu2026sdp} lets a router assign a skill to each state, which conditions a shared diffusion policy, and SOF~\cite{fang2025actionfree} plans over skills learned from action-free video, recovering actions with a separate module.
In both, the transition is implicit.
DexStitch labels the transition as a skill and runs it with the same policy, so it is neither implicit nor a separate module.
The policy is told which skill comes next.

\section{HUMAN-ROBOT DATA CONVERSION}

DexStitch converts egocentric human video into robot-domain observation--action pairs.
Conversion proceeds in three stages (Fig.~\ref{fig:method1}): estimating the hand motion and contact, retargeting it to the robot, and rendering the robot into the frames in place of the human.
The converted video and the retargeted robot trajectory constitute the training data for the policy.
Because conversion is frame-wise, each converted frame keeps the skill label of its source frame.

\begin{figure*}[t]
\centering
\includegraphics[width=0.85\linewidth]{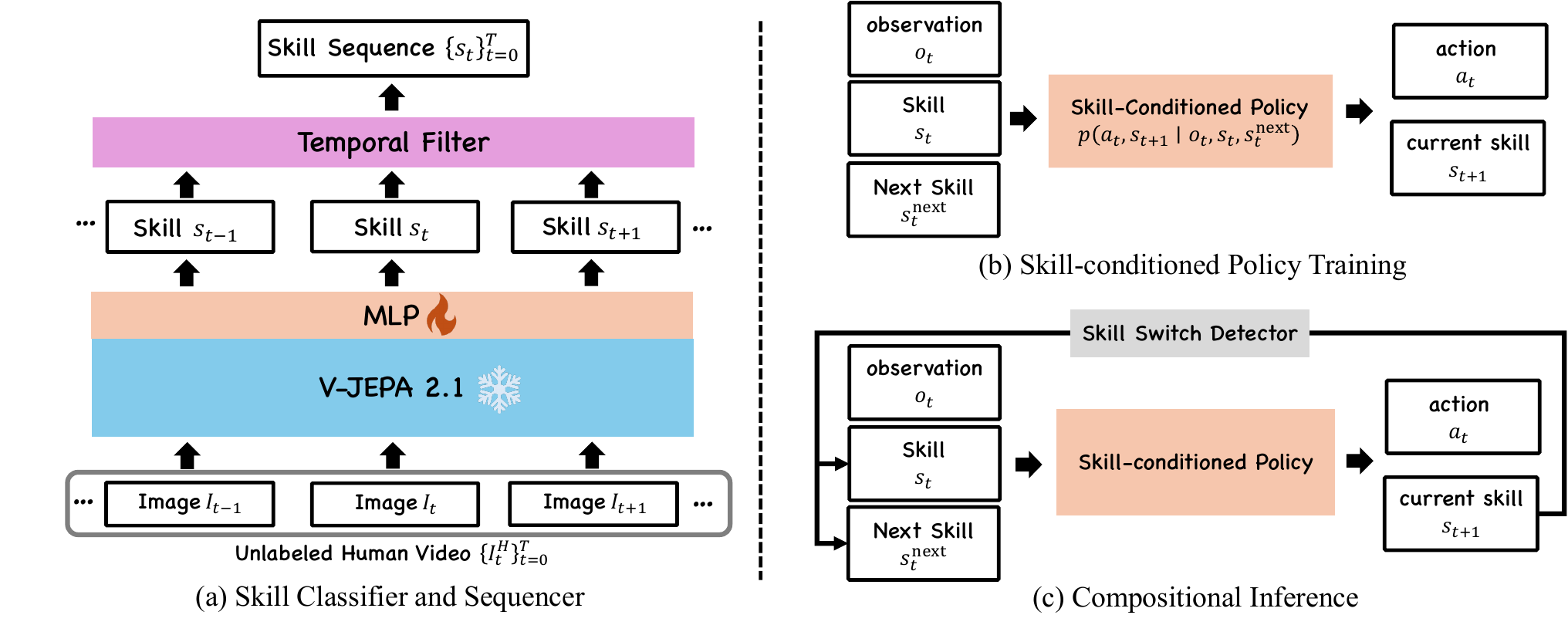}
\caption{
\textbf{Skill Learning and Composition Pipeline.}
(a) A frozen V-JEPA\,2 encoder and a lightweight classification head recover the ordered skill sequence from a human demonstration.
(b) A shared skill-conditioned policy predicts robot actions and skill progression from the current observation and the current--next skill pair.
(c) At inference time, the policy executes the given skill sequence in closed loop, advancing to the next entry according to its predicted skill progression.
}
\label{fig:method2}
\end{figure*}

\subsection{Hand Pose and Contact Estimation}
\label{sec:hand_recon}

We first reconstruct the 3D hand motion with HaWoR~\cite{zhang2025hawor}, which fits the MANO hand model~\cite{romero2017mano} in camera coordinates, giving the hand keypoints $\mathbf{J}_t$ and the wrist pose $\mathbf{T}^{\mathrm{cam}}_{\mathrm{wrist},t}$ at time $t$.
We then estimate human--object contact with HaCo~\cite{jung2025haco}, which predicts a contact probability over the MANO vertices.
Vertex probabilities are pooled by finger and thresholded with hysteresis to obtain temporally stable per-finger contact states.
For each contacting finger, we retain its contacting vertices as $\mathbf{C}_t \subset \mathbb{R}^3$.
These contact points constrain the retargeting in Sec.~\ref{sec:retarget} and the occlusion ordering in Sec.~\ref{sec:contact_render}.

\subsection{Contact-Aware Retargeting}
\label{sec:retarget}

We derive the robot trajectory from the reconstructed hand motion: the keypoints $\mathbf{J}_t$ determine the robot hand articulation, while the wrist pose $\mathbf{T}^{\mathrm{cam}}_{\mathrm{wrist},t}$ specifies the target pose of the robot wrist.
We retarget the hand in two stages.
The first stage applies DexPilot~\cite{handa2019dexpilot}, which maps the MANO keypoints $\mathbf{J}_t$ to the robot hand by matching wrist-to-fingertip vectors.
This provides a kinematic initialization, but the robot fingertips are not guaranteed to reach the demonstrated contact points.
The second stage therefore refines the robot joint angles using the estimated contacts $\mathbf{C}_t$.
For each contacting finger, we minimize the distance from each human contact point to the nearest point on the robot fingertip surface, restricted to surface points whose normal is aligned with the normal at the human contact point.
A quadratic regularizer keeps the solution close to the first stage, and the optimization is solved with L-BFGS-B~\cite{byrd1995limited}, giving the hand configuration $\mathbf{q}^{\mathrm{hand}}_t$.

For the arm, the fixed camera-to-base transform $\mathbf{T}^{\mathrm{cam}}_{\mathrm{base}}$ is first calibrated once for the setup with EasyHeC~\cite{chen2023easyhec}.
We then solve inverse kinematics for $\mathbf{T}^{\mathrm{cam}}_{\mathrm{wrist},t}$ to obtain the arm joint angles $\mathbf{q}^{\mathrm{arm}}_t$, and stack the two: $$\mathbf{x}_t = [\mathbf{q}^{\mathrm{arm}}_t \mid \mathbf{q}^{\mathrm{hand}}_t].$$
Because each frame is processed independently, we smooth the per-frame configurations over time with a median filter followed by a Savitzky--Golay filter.
The result is the robot trajectory $$\mathbf{X}_R = \{\mathbf{x}_t\}_{t=1}^{T},$$ temporally aligned with the human demonstration over its $T$ frames.

\subsection{Inpainting and Compositing}
\label{sec:contact_render}

To produce robot-domain observations aligned with the retargeted trajectory, we remove the human embodiment from each source frame and composite a rendering of the robot while preserving the scene and manipulated objects.
SAM2~\cite{ravi2024sam2} tracks two sets of masks through the video: the human hand and arm, and the manipulated objects.
We inpaint the human mask with ProPainter~\cite{zhou2023propainter}, keeping the object mask protected.
The robot is then rendered at configuration $\mathbf{x}_t$, placed by $\mathbf{T}^{\mathrm{cam}}_{\mathrm{base}}$ and viewed with the intrinsics of the camera that recorded the demonstration.

A direct overlay can place the rendered robot entirely on top of the manipulated object, producing incorrect occlusion around grasping fingers.
To determine which fingers are behind the object, we take the per-finger contact states from Sec.~\ref{sec:hand_recon}.
We assume that when the hand wraps an object, the thumb lies in front of it and the other fingers curl behind, forming an anti-podal grasp.
The non-thumb contacting fingers are therefore the ones to hide: we project their 3D contact points into the image and dilate them into a connected region.
Wherever this region overlaps the object mask, the object is drawn in front of the robot.

Each frame is composited in four layers, from back to front: the inpainted background, the robot fingers the object hides, the object taken from the source frame, and the remaining robot pixels.
The object layer is copied from the source rather than the inpainted plate, since inpainting alters exactly the region where the hand met the object.
Applying this procedure to all frames produces the converted visual sequence $\{\tilde{I}_t^R\}_{t=1}^{T}$, where the tilde marks a quantity converted from the human video and the superscript its robot domain.
Paired with the retargeted robot trajectory $\mathbf{X}_R$, these frames provide robot observations $\tilde{o}_t^R$ and action targets $\tilde{a}_t^R$, yielding $$\tilde{\tau} = \{(\tilde{o}_t^R,\tilde{a}_t^R)\}_{t=1}^{T}.$$

\section{SKILL LEARNING AND COMPOSITION}

We learn to imitate human-demonstrated skills on the robot using the synthetic robot data of Sec.~\ref{sec:contact_render} together with a small set of real robot demonstrations.
Our framework consists of a skill classifier and a skill-conditioned policy, which are jointly used at inference time to execute and compose the demonstrated skills.

\subsection{Skill Classifier}
\label{sec:skill_classifier}

We train a skill classifier to recover the sequence of skills from a human demonstration.
The skill set includes both manipulation behaviors and the transition behavior connecting them, with transition treated as an ordinary skill label.

\subsubsection{Architecture Design and Training}

We keep the pretrained V-JEPA\,2 video encoder~\cite{assran2025vjepa} frozen and attach a small MLP head to its representation.
The encoder emits one token per pair of frames sampled at 4\,FPS, so the head sees the video at 2\,Hz; it reads a causal window of the eight most recent tokens, four seconds of context, and predicts the skill at the latest one.
Freezing the encoder leaves only 296K trainable parameters, which is what allows the classifier to be fitted from the handful of annotated sessions available.

Given a long horizon human video, the classifier is trained to predict a skill label at each token,
$$\hat{s}_t^H = C_\phi(I_t^H),$$
and each token's label is held across the frames it was computed from.

\subsubsection{Skill Annotation Generation}
With the trained skill classifier, we annotate the human video with the skill labels.
We represent a human demonstration as a video $\{I_t^H\}_{t=1}^{T}$ and annotate per-frame skill labels $\{s_t\}_{t=1}^{T}$, where $s_t\in\mathbf{S}$.
We smoothen the frame-wise sequence with temporal filters.
This consolidates consecutive predictions of the same label into an ordered sequence of K skill segments
$$q=(s'_1,...,s'_K).$$
The generated skill labels are then combined with the synthetic robot data to train the robot-executable policy.

\subsection{Skill-Conditioned Policy}
\label{sec:policy}

To reproduce the skills observed in human demonstrations, we learn a single policy that maps each demonstrated skill to the corresponding robot behavior.
The synthetic robot data of Sec.~\ref{sec:contact_render} provides robot observations and actions together with their corresponding human skill labels,
$$\mathbf{D}_{\mathrm{syn}} = \{(\tilde{o}_t^R,\tilde{a}_t^R,s_t)\}_{t=1}^{T},$$
from which the policy learns both individual skills and the motions connecting them.

\subsubsection{Architecture Design}

We condition a shared policy on the current and subsequent skills to support their sequential execution.
Given the current robot observation $o_t$, current skill $s_t$, and next skill $s_t^{\mathrm{next}}$, the policy predicts an action chunk of $H$ steps together with its skill progression:
$$\pi_\theta
\left(
a_{t:t+H},\hat{s}_{t:t+H}
\mid
o_t,s_t,s_t^{\mathrm{next}}
\right).$$
The current skill specifies the behavior to execute, while the next skill provides context for how the behavior should terminate and connect to what follows.
The predicted labels $\hat{s}_{t:t+H}$ capture the progression of the execution and are used for closed-loop skill switching during composition.
We instantiate $\pi_\theta$ with Diffusion Policy~\cite{chi2023diffusion}, where $o_t$ consists of the robot camera observation and the current arm and hand configuration.

\subsubsection{Training}

We first train the policy exclusively on the synthetic robot data $\mathbf{D}_{\mathrm{syn}}$.
This phase lets the policy acquire the full skill repertoire and the inter-skill motions observed in the long-horizon human demonstration.
Because $\mathbf{D}_{\mathrm{syn}}$ covers the complete demonstrated sequence, the policy learns the transitions without any teleoperated demonstration of them.

We then continue training on a mixture of $\mathbf{D}_{\mathrm{syn}}$ and a small set of teleoperated robot demonstrations $\mathbf{D}_{\mathrm{tele}}$.
The real robot data grounds the pretrained policy in the target embodiment and corrects residual discrepancies introduced during conversion, while the synthetic robot data maintains coverage of the human-derived skills and transitions.
This mixed training combines the broad supervision available from human videos with the execution fidelity of limited real-robot data.

\subsection{Compositional Inference}
\label{sec:composition}

Given an new ordered sequence of skills whether by a human demonstration video or script, the robot executes the entries of skills $q$ in order using the skill-conditioned policy.
If a human demonstration is given, the skill classifier first produces an ordered sequence $q=(s'_1,\ldots,s'_K)$, equivalent to the ordered sequence given by script.
While executing $s'_k$, the policy is conditioned on the current and next entries $(s'_k,s'_{k+1})$ and predicts both robot actions and skill labels.

Execution advances through $q$ using the policy's skill predictions rather than a fixed duration.
When the policy consistently predicts the next skill $s'_{k+1}$, the current skill is considered complete and the policy conditions are advanced to $(s'_{k+1},s'_{k+2})$.
We aggregate skill predictions over consecutive time steps to prevent isolated predictions from triggering premature switches.
Repeating this process allows the robot to follow the skill composition extracted from the human demonstration in closed loop.

\begin{figure*}[t]
\centering
\includegraphics[width=0.77\linewidth]{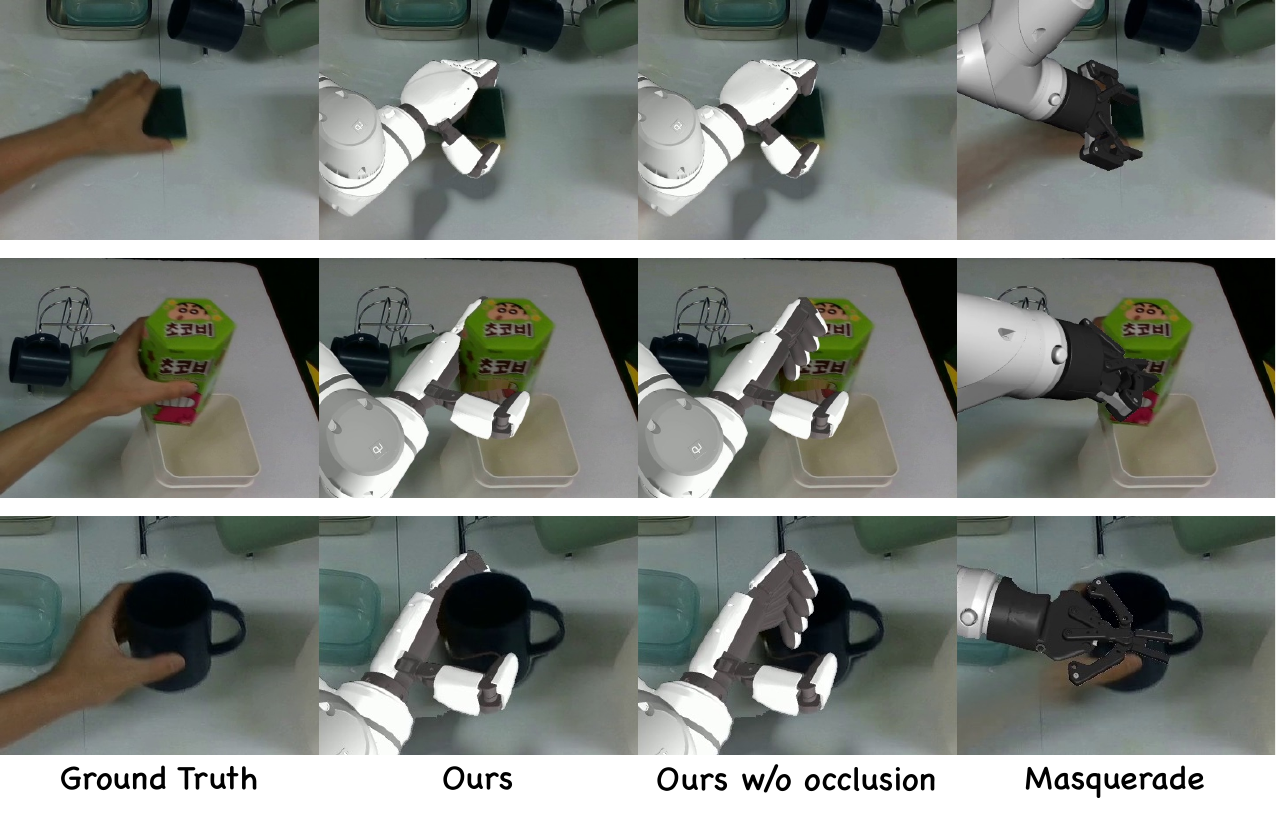}
\caption{\textbf{Synthetic robot observations at the moment of grasp.}
Three grasps from one demonstration: holding a sponge (top), lifting a carton (middle), and hanging a mug on the rack (bottom).
Ours generates realistic robot demonstration images, preserving the hand--object occlusion that the other two lose.}
\label{fig:qualitative}
\end{figure*}

\section{EXPERIMENTS}

\subsection{Experimental Setup}

\paragraph{Hardware}
We conduct all real-robot experiments using an RB5-850E robot arm equipped with an xHand dexterous hand.
Visual observations are captured with an Intel RealSense D435i, of which only the RGB stream is used.

\paragraph{Tasks}
We consider four household manipulation skills: picking and placing a green box, picking and placing a red box, handing over a cup, and stacking bowls.

\paragraph{Dataset}
Our dataset consists of approximately $240$ long-horizon human demonstration episodes in which the four skills are performed in varying orders.
Our pipeline converts these into robot observation--action trajectories.
We additionally collect a small set of teleoperated robot demonstrations---about $8$ per skill ($\sim$32 in total), each executing a single skill in isolation---which are used to finetune the policy.

\paragraph{Implementation}
Our policy is a diffusion policy~\cite{chi2023diffusion}. Visual features are extracted by a ResNet-18
encoder (ImageNet-pretrained), and the action
sequence is denoised by a 1D convolutional U-Net
conditioned on the diffusion timestep and the observation. The observation at
each step is a $224\times224$ RGB image from the camera together with a
$33$-D proprioceptive-and-skill state: the wrist pose ($3$-D position and $6$-D
rotation), $12$ hand joint angles, and the current- and next-skill one-hot
vectors ($6+6$). The policy predicts a $33$-D target of the same layout; at
execution the $6$-DoF arm command is obtained by inverse kinematics from the
predicted wrist pose and the $12$ hand joints are commanded directly, yielding an
$18$-DoF robot action. We use an observation history of $2$ steps and a
prediction horizon of $32$, of which the first $16$ actions are executed, and run
the policy at $20$\,Hz. The forward process uses a DDPM schedule; at inference we denoise with
$10$ DDIM steps.

We train with Adam ($\beta=(0.95,0.999)$, weight decay $10^{-6}$), a cosine
learning-rate schedule, batch size $32$, and mixed precision on a single NVIDIA
A100. The policy is first pretrained for $120$k steps on the converted
human$\rightarrow$robot trajectories ($240$ episodes) at a
learning rate of $10^{-4}$, then finetuned for $70$k steps at $2.5\times10^{-5}$.
For finetuning we mix the $32$ teleoperated episodes, so that teleoperated
data constitutes various ratio(50\%,75\%,90\%,100\%) of the finetuning frames.

\paragraph{Baselines}
We compare against two baselines targeting complementary parts of our system.
For human-to-robot data conversion, we compare with Masquerade~\cite{lepert2025masquerade}.
For reusable skill learning and composition, we compare with UniSkill~\cite{kim2025uniskill}. For a controlled comparison, UniSkill uses the same $120$k-step pretraining and $70$k-step finetuning on the same converted human and teleoperated data as ours, differing only in the skill representation.

\subsection{Skill Learning and Composition}

\begin{table*}[t]
\centering
\caption{
\textbf{Skill Learning and Composition.}
Success rate (\%) for each skill, evaluated both in isolation
(single-skill, 5 trials each) and within four-skill sequences.
Cup: hanging a cup; Stack: stacking bowls.
}
\label{tab:skill_composition}
\small
\setlength{\tabcolsep}{6pt}
\begin{tabular}{lcccccccccc}
\toprule
& \multicolumn{5}{c}{Single-skill} & \multicolumn{5}{c}{Sequence} \\
\cmidrule(lr){2-6}\cmidrule(lr){7-11}
Method & Green & Red & Cup & Stack & Avg. & Green & Red & Cup & Stack & Total \\
\midrule
UniSkill~\cite{kim2025uniskill} & 80 & 20 & 20 & 60 & 45.0 & 0 & 0 & 0 & 0 & 0.0 \\
Ours (DexStitch) & \textbf{100} & \textbf{100} & \textbf{40} & \textbf{80} & \textbf{80.0} & \textbf{80} & \textbf{70} & \textbf{80} & \textbf{80} & \textbf{77.5} \\
\bottomrule
\end{tabular}
\end{table*}

\begin{figure}[t]
\centering
\includegraphics[width=\linewidth]{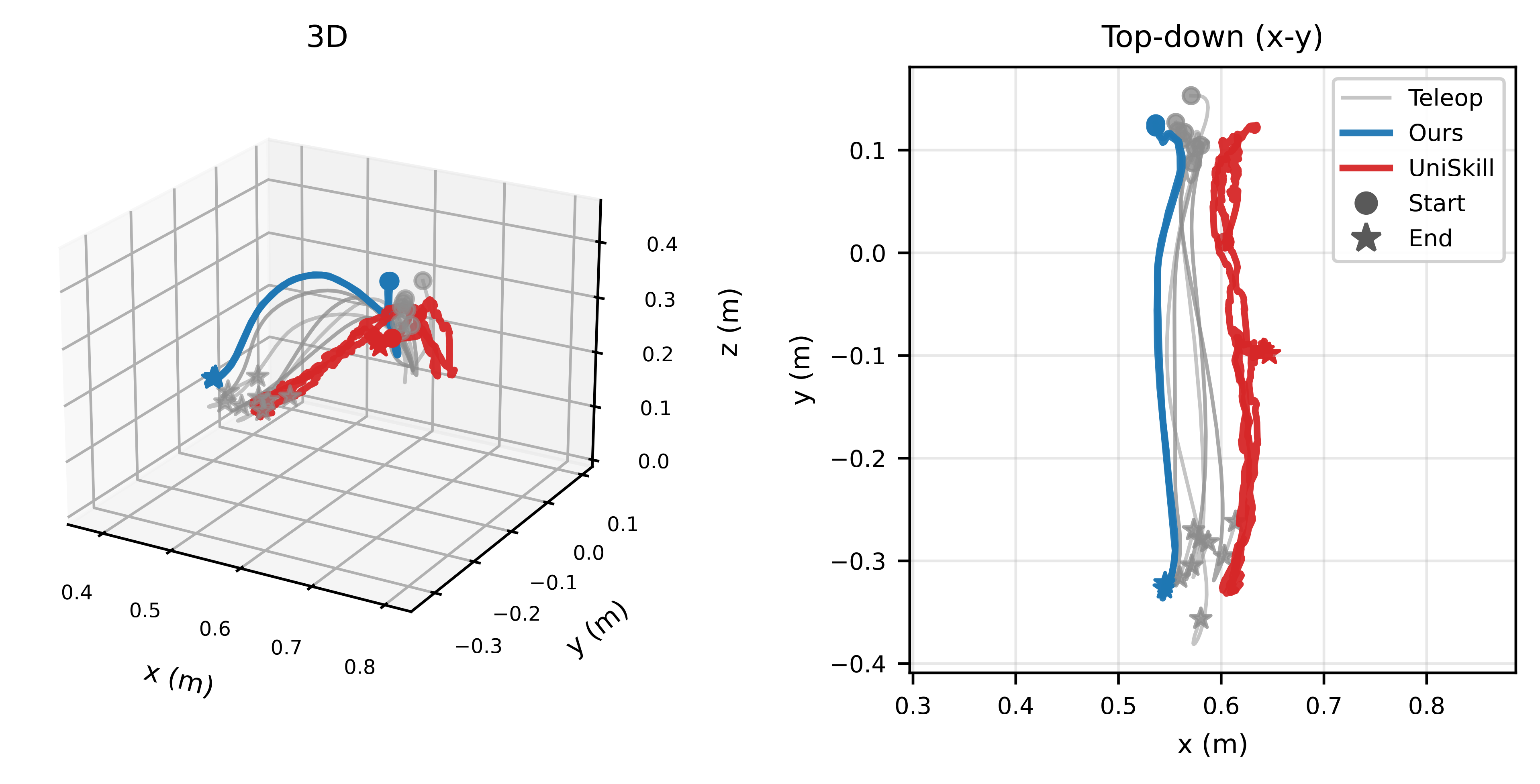}
\caption{
\textbf{End-effector trajectories for the redbox skill.}
Robot base-frame trajectories (m) of the teleoperated demonstrations (gray),
our policy (blue), and UniSkill (red), shown in 3D and top-down ($x$--$y$) views.
}
\label{fig:redbox_traj}
\end{figure}

We evaluate the learned skills in two settings against UniSkill~\cite{kim2025uniskill} under the same skill set (Table~\ref{tab:skill_composition}): a \emph{single-skill} setting, in which each skill is executed on its own for $5$ rollouts, and a \emph{sequence} setting, consisting of $10$ four-skill sequence rollouts (i.e., $40$ skill executions, $10$ per skill).

On single skills, UniSkill performs moderately (\textbf{45.0\%} average), showing that its skill representation captures the individual behaviors.
However, UniSkill provides no explicit mechanism for the transitions between skills, and at the sequence level---where the policy must move from one skill to the next---its success collapses to \textbf{0\%}.
DexStitch instead models transitions through an explicit transition skill together with next-skill conditioning, and stays reliable both on single skills (\textbf{80.0\%}) and in sequences (\textbf{77.5\%}).
Figure~\ref{fig:redbox_traj} illustrates this for the RedBox skill: DexStitch's end-effector trajectories closely follow the teleoperated demonstrations, whereas UniSkill's, though broadly similar in shape, remain offset from the demonstrated motion and do not complete the grasp.
We attribute this gap to the skill representation rather than to data or training budget: both methods use the same synthetic robot data, the same $120$k-step pretraining, and the same teleoperated finetuning, and differ only in how skills are represented---in particular, whether the transitions between skills are modeled.

\subsection{Conversion Quality}

We qualitatively compare our synthetic robot observations with Masquerade~\cite{lepert2025masquerade} and a variant of DexStitch without contact-guided occlusion handling, on the same source demonstrations and with particular emphasis on regions where the robot hand interacts with manipulated objects.
Figure~\ref{fig:qualitative} shows three representative grasps---holding a sponge, lifting a snack carton, and hanging a mug---at the moment of contact.

Masquerade's parallel-jaw gripper cannot express the demonstrated dexterous grasps and composites the gripper over the object it holds, so the resulting images preserve neither the intended hand configuration nor the hand--object occlusion.
Removing contact-guided occlusion handling retains the retargeted dexterous hand but drops the occlusion reasoning: the hand is drawn over the object, and the grasp reads as hovering rather than closing.
Our full method both reproduces the demonstrated finger configuration and renders the object over the fingers whose contact points fall inside its mask, yielding images that read as a robot actually performing the demonstrated grasp.
These comparisons indicate that contact provides a useful cue for preserving interaction consistency when converting human demonstrations to the robot embodiment, particularly around hand--object interactions where human and robot geometry differ.

\begin{table}[t]
\centering
\caption{
\textbf{Skill classification on the held-out session.}
All methods answer the same closed-set question under an identical protocol
(see text)
}
\label{tab:skill_classifier}
\small
\setlength{\tabcolsep}{2.5pt}
\begin{tabular}{lcccc}
\toprule
Method & Acc. $\uparrow$ & Macro-F1 $\uparrow$ & Edit $\uparrow$ & F1@50 $\uparrow$ \\
\midrule
Qwen2.5-VL-7B (zero-shot) & 65.8 & 66.0 & 70.3 & 41.0 \\
InternVL3-8B (zero-shot)  & 57.8 & 56.7 & 47.5 & 28.5 \\
V-JEPA\,2 + MLP (Ours)    & \textbf{92.4} & \textbf{92.7} & \textbf{99.6} & \textbf{99.0} \\
\bottomrule
\end{tabular}
\end{table}

\begin{figure}[t]
\centering
\includegraphics[width=\linewidth]{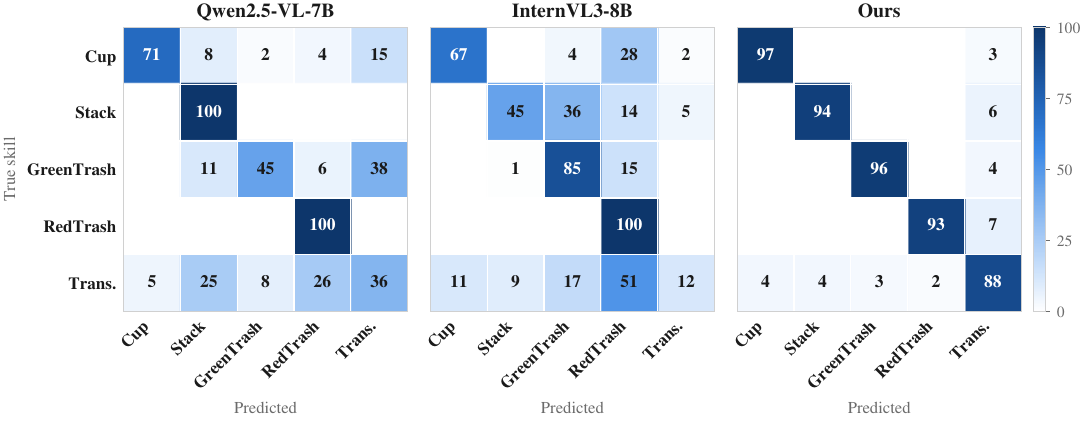}
\caption{
\textbf{Confusion Matrix of Skill Classifier.}
}
\label{fig:skill_confusion}
\end{figure}

\subsection{Skill Classifier Evaluation}

We compare our classifier against Qwen2.5-VL-7B~\cite{bai2025qwen25vl} and InternVL3-8B~\cite{zhu2025internvl3} prompted zero-shot with the same skill set.
Both baselines are queried at the same 2\,Hz decision times from the same causal window, and all three are scored by the same code on the same frames. The VLM baselines see eight $256$-token frames and none abstain.
Of the $240$ human episodes, $216$ are used to train the classifier and the remaining $24$ for evaluation ($18{,}792$ frames).
Besides accuracy and macro-F1 we report segmentation quality, since the classifier exists to cut demonstrations into segments rather than to label isolated moments: edit score compares the predicted skill sequence while ignoring extent, and F1@50 requires at least half overlap with the ground-truth segment.

Table~\ref{tab:skill_classifier} shows the gap widening from roughly thirty points on frame accuracy to a factor of two to three on F1@50.
The baselines label over half the frames correctly but cannot hold one label across a skill, so their predictions fragment and the boundaries do not survive an overlap test; ours recovers 242 segments against 241 in the ground truth, matching the skill order in every episode but one.
The failure concentrates on transitions (F1 \textbf{0.89} ours, \textbf{0.41} Qwen, \textbf{0.18} InternVL, Fig.~\ref{fig:skill_confusion}): with an object in hand the baselines commit to a manipulation skill rather than recognizing motion between skills.
Indeed, the baselines classify the manipulation skills far more accurately than the transitions (Fig.~\ref{fig:skill_confusion}), so what they lack is not visual recognition of the skills but a judgement about temporal state.
Since compositional inference consumes segment boundaries rather than individual frame labels, this margin is what makes the extracted sequences usable.

\begin{table}[t]
\centering
\caption{
\textbf{Ablation Study.}
Overall task success (\%) in the four-skill \emph{sequence} setting.
Ours uses a $90\%$ teleoperation ratio with next-skill conditioning.
}
\label{tab:ablation}
\small
\setlength{\tabcolsep}{6pt}
\begin{tabular}{lc}
\toprule
Variant & Task Success \\
\midrule
\multicolumn{2}{l}{\textit{Training data}}\\
w/o Human data  & 35.0 \\
w/o Teleop data & 5.0 \\
\midrule
\multicolumn{2}{l}{\textit{Skill conditioning}}\\
w/o Transition  & 10.0 \\
w/o Next Skill  & 7.5 \\
\midrule
\multicolumn{2}{l}{\textit{Finetuning ratio}}\\
50\% teleop  & 10.0 \\
75\% teleop  & 20.0 \\
100\% teleop & 50.0 \\
\midrule
\textbf{Ours} & \textbf{77.5} \\
\bottomrule
\end{tabular}
\end{table}

\subsection{Ablation Studies}

All ablations are evaluated in the four-skill sequence setting and compared against our full model (Table~\ref{tab:ablation}); every variant uses the same $90\%$ teleoperation finetuning ratio.

\paragraph{Training data}
We ablate the two training data sources while keeping the architecture fixed (Table~\ref{tab:ablation}).
Using only teleoperated data (\emph{w/o Human data}) reaches \textbf{35.0\%}, while using only the synthetic robot data (\emph{w/o Teleop data}) reaches \textbf{5.0\%}; both fall far below the full model's \textbf{77.5\%}.
Neither source alone suffices: the synthetic robot data provides broad skill coverage, while the teleoperated data grounds the behaviors in the robot's own execution.

\paragraph{Skill conditioning}
We ablate the skill conditioning while keeping the training data unchanged (Table~\ref{tab:ablation}).
Removing the explicit transition skill (\emph{w/o Transition}) lowers success to \textbf{10.0\%}, and removing next-skill conditioning (\emph{w/o Next Skill}) to \textbf{7.5\%}.
Both degrade most at transitions, where the same transition must lead to different subsequent behaviors depending on the skill that follows; the explicit transition skill and the next-skill signal give the policy the context to move toward the appropriate next skill.

\paragraph{Finetuning data ratio}
We vary the proportion of teleoperated robot data in the finetuning mixture, using the synthetic robot data as the remainder.
As shown in Table~\ref{tab:ablation}, task success rises from \textbf{10.0\%} at a \textbf{50\%} teleoperation ratio to a peak of \textbf{77.5\%} at \textbf{90\%}, and then drops to \textbf{50.0\%} at \textbf{100\%} (teleoperation only).
The peak at a $90\%$ ratio shows that retaining a small fraction of the synthetic robot data during finetuning outperforms finetuning on teleoperated data alone. We therefore adopt this $90\%$ ratio for DexStitch.
This indicates that the synthetic robot data provides complementary supervision that teleoperated data alone does not, reducing the amount of robot-collected data required for successful execution.

\section{CONCLUSION}

We presented \textbf{DexStitch}, a framework for learning reusable robot
skills from long-horizon human demonstrations.
DexStitch converts long-horizon human demonstrations into robot-domain
observation--action data using contact-aware retargeting and compositing,
and learns manipulation and transition skills with a skill-conditioned
policy.
A skill classifier extracts skill sequences from human video, enabling
the robot to execute skill sequences in closed loop.
Our experiments show that human demonstrations can provide both
executable robot supervision and reusable skill structure.

Two limits remain.
First, the compositing assumes a wrap grasp, with the thumb in front of the
object and the other fingers curled behind it, so pinch and top-down grasps
fall outside that rule; covering a wider grasp taxonomy is left to future
work.
Second, composition is evaluated on a single fixed four-skill order, so
evaluating orderings held out from training would test the compositional
claim directly.

\bibliographystyle{IEEEtran}
\bibliography{IEEEabrv,references}

\end{document}